\documentclass[10pt,twocolumn,letterpaper]{article}

\usepackage[pagenumbers]{iccv}              
\usepackage{algorithm}
\usepackage{algpseudocode}
\usepackage[hang,flushmargin]{footmisc}
\usepackage[normalem]{ulem}
\usepackage{tikz}
\usepackage[accsupp]{axessibility}

\definecolor{iccvblue}{rgb}{0.21,0.49,0.74}
\newcommand{\nbf}[1]{{\noindent \textbf{#1.}}}
\usepackage[pagebackref,breaklinks,colorlinks,allcolors=iccvblue]{hyperref}

\def\paperID{8422} 
\def\confName{ICCV}
\def\confYear{2025}

\title{4DSegStreamer: Streaming 4D Panoptic Segmentation via Dual Threads}

\author{
Ling Liu\textsuperscript{1}\thanks{Equal contribution} \quad
Jun Tian\textsuperscript{1}\footnotemark[1] \quad
Li Yi\textsuperscript{1,2,3}\thanks{Corresponding author} \\
\textsuperscript{1}IIIS, Tsinghua University \\
\textsuperscript{2}Shanghai Qi Zhi Institute \quad
\textsuperscript{3}Shanghai AI Lab\\
\href{https://llada60.github.io/4DSegStreamer/}{https://llada60.github.io/4DSegStreamer}
}

\begin{document}
\maketitle
\begin{abstract}

4D panoptic segmentation in a streaming setting is critical for highly dynamic environments, such as evacuating dense crowds and autonomous driving in complex scenarios, where real-time, fine-grained perception within a constrained time budget is essential. In this paper, we introduce 4DSegStreamer, a novel framework that employs a Dual-Thread System to efficiently process streaming frames. The framework is general and can be seamlessly integrated into existing 3D and 4D segmentation methods to enable real-time capability. It also demonstrates superior robustness compared to existing streaming perception approaches, particularly under high FPS conditions. The system consists of a predictive thread and an inference thread. The predictive thread leverages historical motion and geometric information to extract features and forecast future dynamics. The inference thread ensures timely prediction for incoming frames by aligning with the latest memory and compensating for ego-motion and dynamic object movements. We evaluate 4DSegStreamer on the indoor HOI4D dataset and the outdoor SemanticKITTI and nuScenes datasets. Comprehensive experiments demonstrate the effectiveness of our approach, particularly in accurately predicting dynamic objects in complex scenes. 

\end{abstract}    
\section{Introduction}
\label{sec:intro}
\indent 

\begin{figure}
    \centering
    \includegraphics[width=1\linewidth]{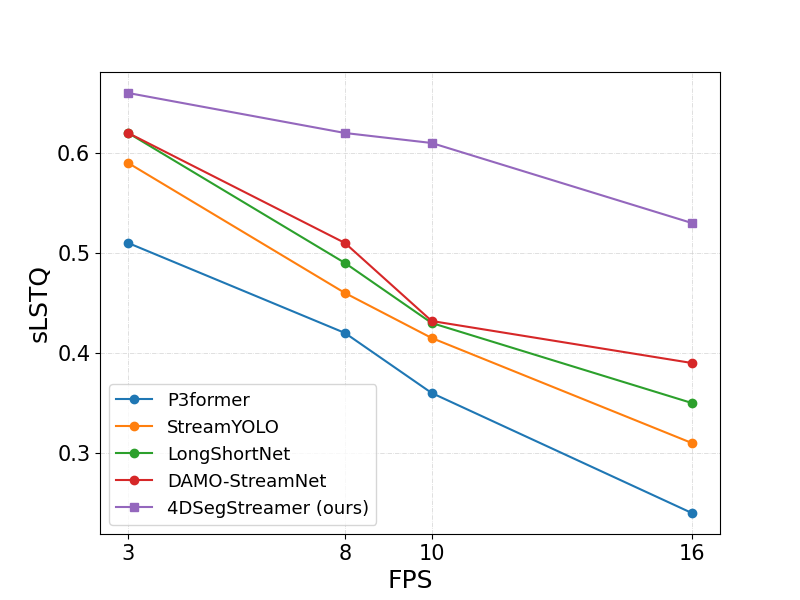}
    \caption{Comparison of streaming performance at different FPS settings on the SemanticKITTI dataset.  Our 4DSegStreamer demonstrates significant performance gains and exhibits a slower performance decline as the FPS increases, indicating its robustness as a more advanced 4D streaming system for panoptic segmentation tasks, particularly in high-FPS scenarios.} 
    \label{fig:teaser}
\end{figure}

Map-free autonomous agents operating in highly dynamic environments require a comprehensive understanding of their surroundings and rapid response capabilities, essential for tasks such as outdoor autonomous driving and indoor robotic manipulation. While low latency may not be critical in static or map-available settings, it becomes a significant challenge in dynamic, map-free environments, where effective navigation and interaction rely on real-time perception. The primary goal of streaming perception is to generate accurate predictions for each incoming frame within a limited time budget, ensuring that perception results remain up-to-date and relevant to the current state of the environment. 

Existing streaming perception research mainly focuses on tasks such as 2D object detection \cite{li2020towards,yang2022real,ghosh2024chanakya,he2023damo,li2023longshortnet,jo2022dade,huang2023mtd,zhang2024transtreaming, ge2024streamtrack, huang2024dyronet}, 2D object tracking \cite{sela2022context, li2023pvt++}, and 3D object detection \cite{han2020streaming,vora2023streaming,frossard2021strobe,abdelfattah2023multi,li2023sodformer,chen2021polarstream} in autonomous driving application, aiming to balance accuracy and latency. However, object bounding boxes are usually insufficient to provide finer-grained knowledge like the object shape or scene context, which is critical for downstream decision-making. For instance, in autonomous driving, relying solely on object detection does not allow the system to accurately identify areas like construction zones or sidewalks, which are essential to avoid for safe navigation.

To achieve a more comprehensive understanding of the scene in a streaming setup, we focus on the challenging task of streaming 4D panoptic segmentation. Given a streaming sequence of point clouds, the goal is to predict panoptic segmentation on each frame within a strict time budget, enabling real-time scene perception. This task is particularly difficult due to the computational overhead and fine-grained perception requirements. Most existing 4D methods \cite{chen2023svqnet, xu2024memory, li2023memoryseg, yilmaz2024mask4former, shi2020spsequencenet, fan2021point, wen2022point, dong2023nsm4d, zhou2021tempnet, wu2024taseg, dewan2020deeptemporalseg, aygun20214d, liu2023mars3d, kreuzberg20224d, zhu20234d} fail to achieve real-time perception and the fluctuations in computing resources introduce additional latency inconsistencies, further complicating streaming 4D panoptic segmentation task.


To address the challenges of real-time dense perception in streaming 4D panoptic segmentation, we introduce 4DSegStreamer, a general system designed to enable existing segmentation methods to operate in real time. 4DSegStreamer utilizes a novel dual-thread system with a predictive thread maintaining geometry and motion memories in the scene and an inference thread facilitating rapid inference at each time step. The key idea behind 4DSegStreamer involves dividing the streaming input into key frames and non-key frames based on the model’s latency. In the predictive thread, we meticulously compute geometric and motion features at key frames and utilize these features to continuously update the memories, enabling long-term spatial-temporal perception. To support efficient memory queries, the memories are also utilized to predict future dynamics, guiding how a future frame can effectively adjust for potential movement when querying the geometry memory. In the inference thread, each incoming frame is first positionally aligned with the current geometry memory by compensating for the forecasted motion. It then swiftly queries the hash table-style memory to obtain per-point labels. The two threads together allow both fast and high-quality streaming 4D panoptic segmentation.

Our contributions to this work can be summarized as: 
\begin{itemize}
    \item We introduce a new task for streaming 4D panoptic segmentation, advancing real-time, fine-grained perception for autonomous systems in dynamic environments.
    \item We propose a novel dual-thread system that includes a predictive thread and an inference thread, which is general and applicable to existing segmentation methods to achieve real-time performance. The predictive thread continuously updates memories by leveraging historical motion and geometric features to forecast future dynamics. The inference thread retrieves relevant features from the memory through geometric alignment with the forecasted motion, using ego-pose transformation and inverse flow iteration.
    \item Through extensive evaluations in outdoor datasets SemanticKITTI and nuScenes, as well as the indoor HOI4D dataset, our system significantly outperforms existing SOTA streaming perception and 4D panoptic segmentation methods. Moreover, our approach demonstrates superior robustness compared to other streaming perception methods shown in Fig.~\ref{fig:teaser}, particularly under high-FPS scenarios. These results highlight the effectiveness and value of our method for 4D streaming segmentation.
\end{itemize}

\begin{figure*}
    \centering
    \includegraphics[width=1\linewidth]{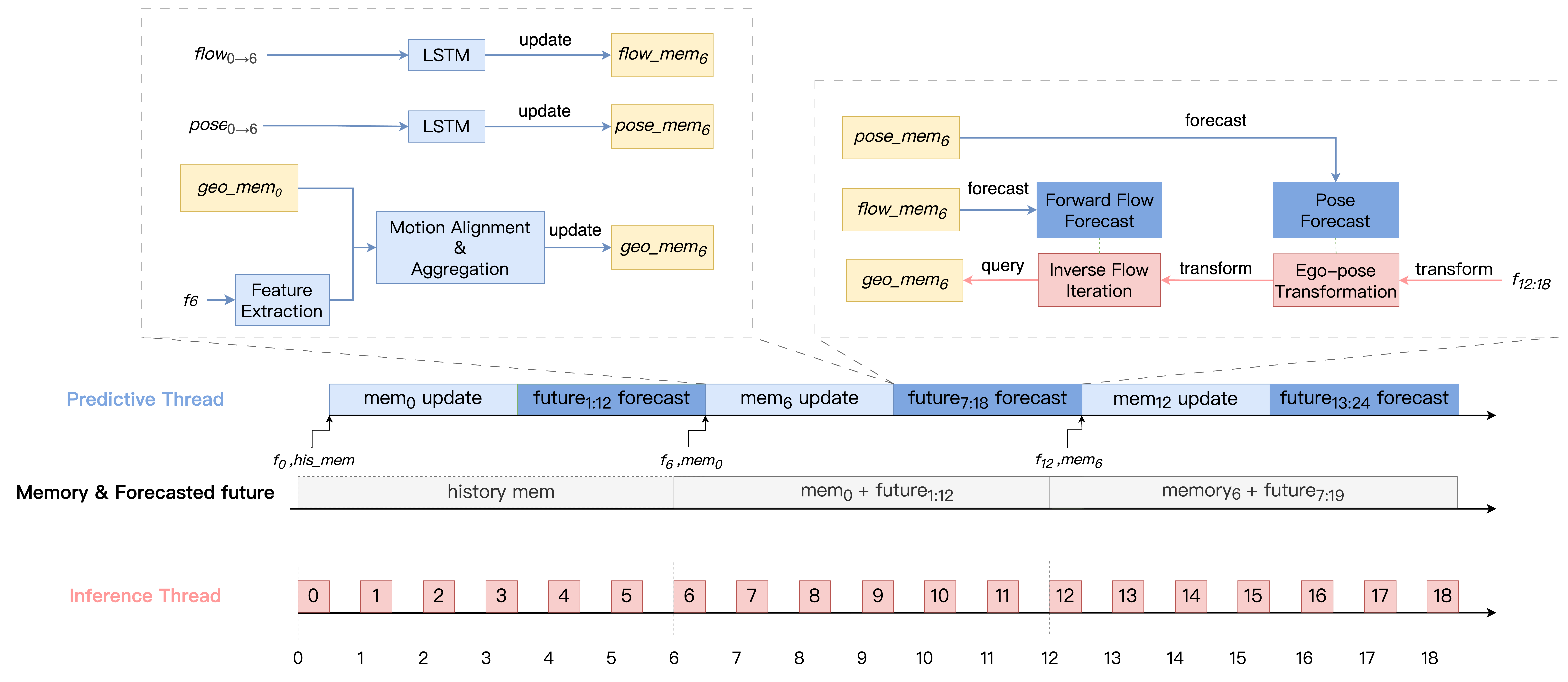} 
    \caption{\textbf{4DSegStreamer}: The dual-thread system consists of a predictive thread and an inference thread, enabling real-time query for unseen future frames. The \textcolor[rgb]{0.49,0.65,0.88}{predictive thread} updates the geometric and motion memories with the latest extracted feature and leverages the historical information to forecast future dynamics. The \textcolor[rgb]{1,0.6,0.6}{inference thread} retrieves per-point predictions by geometrically aligning them with the current memory using ego-pose and dynamic object alignment. Here, mem$_i$ denotes the memory updated with the latest key frame f$_i$, while f${_{i:j}}$ represents incoming frame${_{i, i+1, ..., j}}$.}
    \label{fig:mainpic}
\end{figure*}

\section{Related Work}
\label{sec:related}

\subsection{Streaming Perception}
\indent 

 In the streaming perception, the inherent challenge lies in predicting results in the future state, in order to minimize the temporal gap between input and output timestep. Most previous studies concentrate on developing forecasting modules specifically tailored for this streaming setting. Stream \cite{li2020towards} firstly introduces the streaming setting and utilizes the Kalman Filters to predict future bounding boxes. StreamYOLO \cite{yang2022real} designs a dual-flow perception module, which incorporates dynamic and static flows from previous and current features to predict the future state. DAMO-StreamNet \cite{he2023damo} and LongShortNet \cite{li2023longshortnet} leverages spatial-temporal information by extracting long-term temporal motion from previous multi-frames and short-term spatial information from the current frame for future prediction. Different from previous researches which only forecast one frame ahead and thus the prediction output is limited within a single frame, DaDe \cite{jo2022dade} and MTD \cite{huang2023mtd} considering previous prediction time, adaptively choose the corresponding future features. Transtreaming \cite{zhang2024transtreaming} designs an adaptive delay-aware transformer to select the prediction from multi-frames future that best matches future time. 



Several studies have explored streaming perception in LiDAR-based 3D detection  \cite{han2020streaming,vora2023streaming,frossard2021strobe,abdelfattah2023multi,li2023sodformer,chen2021polarstream}. Lidar Stream \cite{han2020streaming} segments full-scan LiDAR points into multiple slices, processing each slice at a higher frequency compared to using the full-scan input. Although ASAP \cite{wang2023we} introduces a benchmark for online streaming 3D detection, it relies on camera-based methods using images as input. 



\subsection{4D Point Cloud Sequence Perception}
\indent 

4D point cloud sequence perception methods integrate temporal consistency and spatial aggregation through advanced memory mechanisms. These methods are generally categorized into voxel-based \cite{chen2023svqnet, xu2024memory, li2023memoryseg,yilmaz2024mask4former} and point-based \cite{shi2020spsequencenet, fan2021point, wen2022point, dong2023nsm4d, zhou2021tempnet, wu2024taseg, dewan2020deeptemporalseg, aygun20214d, liu2023mars3d, kreuzberg20224d, zhu20234d, marcuzzi2023mask, li2024streammos, kreuzberg20224d} approaches.

For the point-based methods, SpSequenceNet \cite{shi2020spsequencenet} aggregates 4D information on both a global and local scale through K-nearest neighbours. NSM4D \cite{dong2023nsm4d} introduces a historical memory mechanism that maintains both geometric and motion features derived from motion flow information, thereby enhancing perception capabilities.
Eq-4D-StOP \cite{zhu20234d} introduces a rotation-equivariant neural network that leverages the rotational symmetry of driving scenarios on the ground plane.


For the voxel-based methods, SVQNet \cite{chen2023svqnet} develops a voxel-adjacent framework that leverages historical knowledge with both local and global context understanding. This work is further optimized by the implementation of hash query mechanisms for computation acceleration, and is further accelerated by hash query mechanisms. MemorySeg \cite{li2023memoryseg} incorporates both point and voxel representations for contextual and fine-grained details learning. 
Mask4Former \cite{yilmaz2024mask4former} introduces a transformer-based approach unifying semantic instance segmentation and 3D point cloud tracking.

\subsection{Fast-slow Dual System Methods}
\indent 
The fast-slow system paradigm, merging efficient lightweight models with powerful large-scale models, has gained attention. For instance, DriveVLM-Dual \cite{tian2024drivevlm} integrates 3D perception and trajectory planning with VLMs for real-time spatial reasoning, while FASIONAD \cite{qian2024fasionad} introduces an adaptive feedback framework for autonomous driving, combining fast and slow thinking to improve adaptability in dynamic environments.

While 4DSegStreamer is not explicitly designed as a fast-slow system, its dual-thread architecture shares some conceptual similarities. 
The predictive thread acts as a slow component, responsible for maintaining memory and forecasting future dynamics, while the inference thread acts as a fast component, enabling real-time inference through efficient feature retrieval. However, unlike traditional fast-slow systems that rely on separate models for fast and slow tasks, 4DSegStreamer integrates both components into a unified pipeline, enabling seamless interaction between memory updates and real-time queries. 


% ...
\section{Streaming 4D Panoptic Segmentation}
\indent 

We propose a new task of streaming 4D panoptic segmentation. Similar to the traditional streaming perception paradigm, streaming 4D panoptic segmentation conducts the panoptic segmentation in an online manner. The key challenge is ensuring that each incoming frame is processed and predicted within an ignorant small time budget, even if the processing of the current frame is not complete. Our goal is to develop an approach that finds a trade-off between accuracy and efficiency to enable real-time inference for the Streaming 4D Panoptic Segmentation task.

\label{sec:preliminary}

\section{Method}
\indent 

In this section, we introduce 4DSegStreamer (see Fig.~\ref{fig:mainpic}) to address the challenges of streaming 4D panoptic segmentation. The key idea is to divide the streaming frames into key frames and non-key frames, where geometric and motion features are continuously extracted at key frames to update the memories, and subsequently used to accelerate inference for each future frame. 4DSegStreamer employs a novel dual-thread system comprising a predictive thread and an inference thread, which is general and can be applied to various segmentation methods to enable their real-time performance. The system contains three key stages, including memory update to maintain spatial-temporal information of geometric and motion features, ego-pose future alignment to cancel ego-motion, and dynamic object future alignment to eliminate dynamic object movement.

\subsection{Dual-thread system}
\indent 

Unlike previous works in 2D streaming perception, which focus on object detection and tracking by predicting the transformation of bounding boxes, 4D panoptic segmentation must establish correspondences between past predictions and unseen future point clouds across multiple frames due to the latency. To address this challenge, we simplify the real-time inference problem using a dual-thread system. This system consists of a Predictive Thread for memory updating and future dynamics forecasting and an Inference Thread that allows incoming future points to quickly retrieve the corresponding features from memory, ensuring efficient inference within the limited time constraints.

\nbf{Predictive thread} We continuously update the geometric and motion memories with the latest available frame as a key frame. Leveraging the spatial-temporal information in the motion memories, we forecast the future camera and dynamic object movement to align future frames with corresponding features in geometric memory, thereby accelerating the inference in the inference thread.

\begin{figure}[t]
    \centering
    \includegraphics[width=1\linewidth]{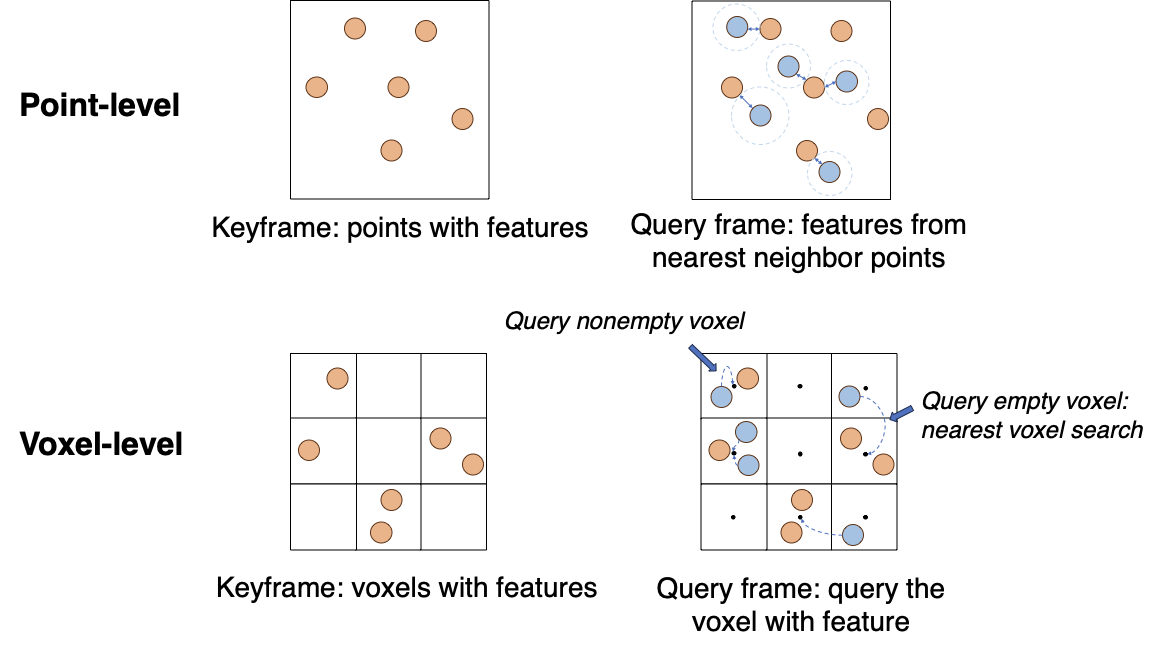}
    \caption{Point-level and voxel-level methods in inference thread: \textcolor{orange}{orange} points indicate the extracted features corresponding keyframe points, while \textcolor[rgb]{0.25, 0.5, 0.75}{blue} points indicate the aligned incoming frame points querying the features from memory.}
    \label{fig:pointvsvoxel}
\end{figure}

\nbf{Inference thread} Each incoming frame is geometrically aligned with the latest memory using forecasted pose and flow. The corresponding features are then retrieved from the geometric memory using two query strategies, as illustrated in Fig.~\ref{fig:pointvsvoxel}. In our approach, we use a hash table-style memory that allows direct access to corresponding voxel features via their indices and apply nearest neighbor search only for points querying empty voxels. These retrieved features are subsequently passed through a lightweight prediction head to produce the final output.

The dual-thread system operates in parallel and shares the memory to process streaming point clouds in real time. The overall inference latency is primarily determined by the inference thread, which is lightweight and fast, while the predictive thread maintains long-term spatio-temporal memories by continuously updating them with the latest features. At each timestamp, the inference thread retrieves relevant features from memory through motion alignment, ensuring real-time inference.

\subsection{Geometric Memory Update}
\indent 

Our system is general and can be integrated into both 3D and 4D segmentation backbones, where features are stored at the voxel level for fast query in the inference thread and aggregated to update using the latest keyframe via motion alignment. The memory system leverages a sparse variant of ConvGRU \cite{ballas2015delving, li2023memoryseg} to perform geometric memory updates efficiently.

Upon the arrival of a keyframe, we first perform motion alignment by transforming the previous memory state $h_{t-k}$ to the current frame, resulting in the aligned memory $h_{t-k}^{\prime}$:

\begin{equation}
    h_{t-k}^{\prime} = f_{t-k \rightarrow t} \left( p_{t-k \rightarrow t} \cdot h_{t-k} \right)
\end{equation}

where $p_{t-k \rightarrow t}$ denotes ego-pose transformation and $f_{t-k \rightarrow t}$ represents dynamic object flow transformation. Both transformation are applied to convert the memory coordinates into the current keyframe's coordinate space, aligning both static and dynamic objects. 

Subsequently, the geometric memory is updated using the current frame's feature embeddings $f_t$:

\begin{equation}
    \begin{split}
        z_t &= \sigma(\Psi_z(f_t, h_{t-k}^{\prime})),\\
        r_t &= \sigma(\Psi_r(f_t, h_{t-k}^{\prime})),\\
        \hat{h}_t &= tanh(\Psi_u(f_t, r_t, h_{t-k}^{\prime})),\\
        h_t &= \hat{h}_t\cdot z_t + \hat{h}_{t-k}\cdot (1-z_t),
    \end{split}
\end{equation}

where $\Psi_r, \Psi_z, \Psi_u$ are sparse 3D convolution blocks. $z_t$ and $r_t$ are activation gate and reset gate to update and reset the memory. The updated memory retains the latest spatial-temporal information to support future dynamics forecasting and efficient feature queries.

\begin{figure}[t]
    \centering
    \includegraphics[width=1\linewidth]{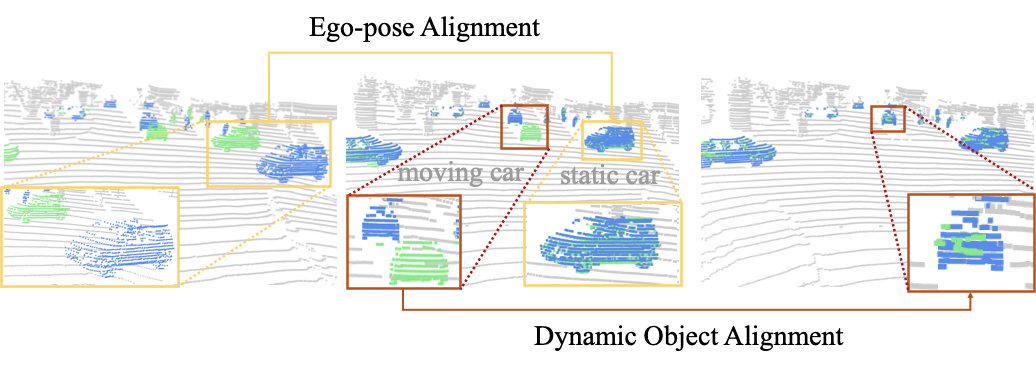}
    \caption{Ego-pose Alignment and Dynamic Object Alignment: The \textcolor[rgb]{0.56,0.93,0.56}{green} points represent the previously processed frame that has been used to update the memories and the \textcolor[rgb]{0.39,0.58,0.92}{blue} points are the current querying frame. The \textcolor[rgb]{1, 0.83, 0.38}{yellow} box highlights static objects that can be aligned through ego-pose alignment. The \textcolor[rgb]{0.73,0.31, 0.1}{red} box indicates dynamic objects, which require dynamic object alignment to achieve proper alignment.}
    \label{fig:alignment}
\end{figure}


\subsection{Ego-pose Future Alignment}
\label{sec:ego-pose alignment}
\indent 



As seen in Fig.~\ref{fig:alignment}, the static car in the incoming frame is positioned differently from the same car in memory. To ensure temporal consistency in dynamic environments, we utilize ego-pose forecasting to compensate for camera motion and align the current memory with future frames.

In many outdoor applications, such as autonomous driving, ego-pose information is typically available from onboard sensors. However, in indoor scenarios, such as an embodied robot operating in a room, obtaining pose information is often challenging and requires pose estimation.

Depending on whether the camera pose is available, we define two settings:

\begin{itemize}
    \item \textit{Known pose setting}: we directly use the relative pose to align future frames with the feature memory coordinates.
    \item  \textit{Unknown pose setting}: we utilize the pose estimated by Suma++ \cite{chen2019suma++} between key frames to update the ego-motion memory, and then use the ego-pose forecaster to propagate the future ego-pose motion, ensuring proper alignment and eliminating ego motion.
\end{itemize}

Here we introduce the unknown pose setting. When a keyframe $x_{t}$ is coming, the estimator $E$ will estimate the relative ego motion between last keyframe $x_{t-k}$ and current keyframe $x_{t}$:

\begin{equation}
p_{t-k\rightarrow t} = E(x_{t-k}, x_{t})
\end{equation}

Then, utilize the key pose to update the ego-pose memory $memp_{t-k}$, we have:
\begin{equation}
    memp_{t} = W(p_{t-k\rightarrow t}, memp_{t-k})
\end{equation}

where $W$ indicates the memory update function which we use the LSTM \cite{graves2012long}. In order to forecast the relative pose m frames ahead for the future frame $x_{t+m}$ using pose forecaster $F$, we have:
\begin{equation}
    p_{t\rightarrow t+m} = F(memp_t, m)
\end{equation}

where the ego-pose forecaster is designed in a multi-head structure, with each head predicting the future pose for a fixed number of frames ahead.

\subsection{Dynamic Object Future Alignment}
\indent 

Compared to static objects, dynamic objects exhibit both ego-motion and independent self-movement, with varying velocities and directions, as seen in the moving car in Fig.~\ref{fig:alignment}. To achieve fine-grained self-motion alignment for dynamic objects and fast query, we introduce the Future Flow Forecasting in the predictive thread and the Inverse Forward Flow in the inference thread.

\nbf{Future Flow Forecasting} During training, we use FastNSF \cite{li2023fast} to obtain supervised ground truth flows. In inference time, the process is similar to ego-pose future alignment in Sec \ref{sec:ego-pose alignment}. We utilize zeroFlow \cite{vedder2023zeroflow}, a lightweight model distilled from FastNSF, to estimate key flows between keyframes. These key flows are then input into the LSTM \cite{graves2012long} to forecast future flows, supporting the fast alignment of dynamic objects across memory and incoming frames. 

\nbf{Inverse Forward Flow Iteration} To enable efficient feature querying during inference, we leverage forecasted forward flows to align the geometric memory with future frames. However, directly applying forward flows to the memory is time-consuming for the predictive thread, as it requires constructing a new nearest-neighbor tree at each future timestamp to enable fast access to the geometric memory. Although backward flow is more efficient that it maps incoming points to the pre-built nearest-neighbor tree of the latest memory, directly forecasting backward flow is challenging due to the unknown number and positions of future points, which leads to degraded performance (see Tab.~\ref{tab:ab_flow}).

To balance the efficiency and accuracy, we propose the Inverse Forward Flow Iteration. The goal of our method is to find the corresponding point $x$ in history memory with the current query point $y$. The correspondence satisfies:
\begin{equation}
    g(x) = x = y - flow(x)
\end{equation}
where flow(x) indicates the forecasted forward flow at point x, and -flow(x) represents the inverse forward flow.

Then we want to find a fixed point $x^*$ such that $x^*=g(x^*)$. Given an initial guess $x_0=y$, define the iteration as:
\begin{equation}
    x_{n+1} = g(x_n) = y-flow(x_n)
\end{equation}

The sequence $\{x_n\}$ will converge to the fixed point $x^*$ if g(x) is a contraction mapping, i.e., if there exists a constant $L\le 1$, such that for all $x_1$ and $x_2$ satisfy:
\begin{equation}
    \label{eq:converge}
    |g(x_1)-g(x_2)|\leq L|x_1-x_2|
\end{equation}

The stopping iteration condition is
\begin{equation}
    |x_{n+1} - x_n|\le \epsilon
\end{equation}
where $\epsilon$ is the predefined tolerance, indicating $x_n$ has converged to the solution. To hold this condition, we need $g(x)$ to be Lipschitz continuous, and its Lipschitz constant $L \le 1$. Thus, we assume $\left|flow^{\prime}\left(x\right)\right| \leq 1$ for each differentiable point $x$. The detailed proof is provided in Supp.~\ref{supp:proof}.

The query point iteratively finds the local forecasted forward flow in memory, then backtracks through the inverse of this forward flow. The process continues until the distance between current query position $p^{\prime}$ and the point $p$ closely approximates the inverse of the forward flow. The pseudo-code for this process is as follows:

\begin{algorithm}[H]
\caption{Iterative Inverse Forward Flow Method}
\label{alg:inverse-forward-flow}
\begin{algorithmic}[1]
\Require forecast forward flow query $Q$, stop threshold $\epsilon$, maximum iterations $N_{max}$
\For{each point $p$ in the non-key frame}
    \State Initialize current query position $p' \gets p$
    \State Initialize iteration counter $n \gets 0$
    \State Inverse(f) $\gets -f$
    \While{$\| (p'-p) + Q(p') \| \geq \epsilon$ \textbf{and} $n < N_{max}$}
        \State Query local forecast forward flow $f \gets Q(p')$
        \State Update track position: $p' \gets p + \text{Inverse}(f)$
        \State Increment iteration counter: $n \gets n + 1$
    \EndWhile
\EndFor
\end{algorithmic}
\end{algorithm}

\label{sec:method}

\begin{table*}
\caption{SemanticKITTI validation set result in \textit{unknown pose} streaming setting. The best is highlighted in \textbf{bold}. sX indicates the metric X in the streaming setting. PQ$_d$ and PQ$_s$ refer to the evaluation for dynamic and static points, respectively. PQ$_{th}$ evaluates the thing class and PQ$_{st}$ evaluates the stuff class.}
  \centering
  \begin{tabular}{l|ccc|ccc|cccc}
    \toprule
    Method & sLSTQ & S$_{assoc}$ & S$_{cls}$ & sPQ & sRQ & sSQ & sPQ$_{d}$ & sPQ$_{s}$ & sPQ$_{th}$ & sPQ$_{st}$ \\
    \midrule
    
    StreamYOLO \cite{yang2022real} & 0.415 & 0.321 & 0.536 & 0.373 & 0.478 & 0.664 & 0.429 & 0.371 & 0.388 & 0.364 \\
    LongShortNet \cite{li2023longshortnet} & 0.430 & 0.341 & 0.541 & 0.392 & 0.472 & 0.673 & 0.452 & 0.391 & 0.400 & 0.386 \\
    DAMO-StreamNet \cite{he2023damo} & 0.432 & 0.341 & 0.546 & 0.392 & 0.472 & 0.674 & 0.459 & 0.391 & 0.400 & 0.388  \\
    Mask4Former \cite{yilmaz2024mask4former} & 0.515 & 0.464 & 0.572 & 0.485 & 0.594 & 0.691 & 0.571 & 0.413 & 0.538 & 0.422   \\
    Eq-4D-StOP \cite{zhu20234d} & 0.504 & 0.452 & 0.563 & 0.477 & 0.578 & 0.691 & 0.543 & 0.412 & 0.529 & 0.423  \\
    PTv3 \cite{wu2024point} & 0.536 & 0.492 & 0.586 & 0.567 & 0.612 & 0.704 & 0.638 & 0.464 & 0.575 & 0.459 \\
    4DSegStreamer (P3Former) & 0.613 & 0.627 & 0.599 & 0.602 & 0.679 & 0.723 & 0.711 & 0.479 & 0.625 & 0.481  \\
    4DSegStreamer (Mask4Former) & \textbf{0.688} & \textbf{0.706} & \textbf{0.621} & \textbf{0.634} & \textbf{0.701} & \textbf{0.752} & \textbf{0.744} & \textbf{0.486} & \textbf{0.660} & \textbf{0.497}  \\
    \bottomrule
  \end{tabular}
  \label{tab:sk_wopose}
\end{table*}

\begin{table*}
\caption{SemanticKITTI validation set result in \textit{known pose} streaming setting. The best is highlighted in \textbf{bold}. sX indicates the metric X in the streaming setting. PQ$_d$ and PQ$_s$ refer to the evaluation for dynamic and static points, respectively. PQ$_{th}$ evaluates the thing class and PQ$_{st}$ evaluates the stuff class.}
  \centering
  \begin{tabular}{l|ccc|ccc|cccc}
    \toprule
    Method & sLSTQ & S$_{assoc}$ & S$_{cls}$ & sPQ & sRQ & sSQ & sPQ$_{d}$ & sPQ$_{s}$ & sPQ$_{th}$ & sPQ$_{st}$ \\
    \midrule
    StreamYOLO \cite{yang2022real} & 0.439 & 0.356 & 0.541 & 0.384 & 0.468 & 0.715 & 0.432 & 0.383 & 0.392 & 0.382  \\
    LongShortNet \cite{li2023longshortnet} & 0.446 & 0.360 & 0.553 & 0.412 & 0.489 & 0.719 & 0.459 & 0.410 & 0.413 & 0.399  \\
    DAMO-StreamNet \cite{he2023damo} & 0.446 & 0.362 & 0.551 & 0.425 & 0.489 & 0.724 & 0.460 & 0.412 & 0.414 & 0.401  \\
    Mask4Former \cite{yilmaz2024mask4former} & 0.564 & 0.539 & 0.592 & 0.520 & 0.613 & 0.734 & 0.623 & 0.460 & 0.592 & 0.467   \\
    Eq-4D-StOP \cite{zhu20234d} & 0.557 & 0.530 & 0.585 & 0.520 & 0.619 & 0.732 & 0.625 & 0.459 & 0.594 & 0.465  \\
    4DSegStreamer (P3Former) & 0.655 & 0.703 & 0.610 & 0.687 & 0.774 & 0.816 & 0.782 & 0.560 & 0.704 & 0.531  \\
    4DSegStreamer (Mask4Former) & \textbf{0.701} & \textbf{0.722} & \textbf{0.648} & \textbf{0.704} & \textbf{0.811} & \textbf{0.838} & \textbf{0.803} & \textbf{0.579} & \textbf{0.741} & \textbf{0.552}  \\
    \bottomrule
  \end{tabular}
  \label{tab:sk_wpose}
\end{table*}

\section{Experiments}
\label{sec:experiment}

\indent

We present the experimental setup and benchmark results on two widely used outdoor LiDAR-based panoptic segmentation datasets, SemanticKITTI\cite{behley2019semantickitti} and nuScenes\cite{caesar2020nuscenes}, as well as the indoor dataset HOI4D\cite{liu2022hoi4d}.

\subsection{Settings}

\nbf{SemanticKITTI \cite{behley2019semantickitti}} SemanticKITTI is a large-scale dataset for LiDAR-based panoptic segmentation, containing 23,201 outdoor scene frames at 10 fps. Unlike traditional 4D panoptic segmentation, streaming 4D panoptic segmentation also involves distinguishing between moving and static objects, since the ability to perceive moving objects is significant in streaming perception. This adds 6 additional classes for moving objects (e.g., "moving car") to the standard 19 semantic classes. In total, there are 25 classes, including 14 thing classes and 11 stuff classes.

\nbf{nuScenes \cite{caesar2020nuscenes}} nuScenes is a publicly available autonomous driving dataset with 1,000 scenes captured at 2 fps. We extend the per-point semantic labels to distinguish between moving and non-moving objects using ground truth 3D bounding box attributes. This extension includes 8 moving object classes and 16 static object classes, totaling 18 thing classes and 6 stuff classes. 

\nbf{HOI4D \cite{liu2022hoi4d}} HOI4D is a large-scale egocentric dataset focused on indoor human-object interactions. It contains 3,865 point cloud sequences, with 2,971 for training and 892 for testing. Each sequence has 300 frames captured at 15 fps.

\nbf{Evaluation metrics}
We use PQ and LSTQ in streaming setting (denoted as sPQ and sLSTQ) as our main metrics to evaluate panoptic segmentation performance. Furthermore, we divide the sPQ into four components: $\text{sPQ}_{d}$ for dynamic objects, $\text{sPQ}_{s}$ for static objects, $\text{sPQ}_{th}$ for thing classes, and $\text{sPQ}_{st}$ for stuff classes. In the streaming setting, evaluation of each frame must occur at every input timestamp, according to the dataset’s frame rate. If the computation for the current frame is not completed in time, we use the features from the last completed frame to query the results and perform the evaluation. 



\nbf{Implementation details}
We choose P3Former \cite{xiao2024position} and Mask4Former \cite{yilmaz2024mask4former} as our backbone model, which is originally a SOTA method for 3D and 4D panoptic segmentation. By incorporating the ego pose and flow alignment strategies we proposed, along with memory construction, they can also achieve good performance in 4D streaming panoptic segmentation. We first train the model on each dataset, then freeze it for feature extraction. The remaining components, including ego-pose forecasting, forward flow forecasting, and history memory aggregation, are trained subsequently. For the inverse flow iteration, the maximum iterations patience is set to 10. All models are trained on 4 NVIDIA GTX 3090 GPUs and evaluated on a single NVIDIA GTX 3090 GPU.

\begin{table}
    \caption{nuScenes validation set result in \textit{unknown pose} streaming setting. The best is highlighted in \textbf{bold}.}
  \centering
  \begin{tabular}{l|cccc}
    \toprule
    Method & sLSTQ & sPQ & sPQ$_{d}$ & sPQ$_{s}$ \\
    \midrule
    StreamYOLO \cite{yang2022real}  & 0.596 & 0.581 & 0.569 & 0.591 \\
    LongShortNet \cite{li2023longshortnet} & 0.610 & 0.603 & 0.579 & 0.607 \\
    DAMO-StreamNet \cite{he2023damo}  & 0.623 & 0.607 & 0.601 & 0.612 \\
    Mask4Former \cite{yilmaz2024mask4former}  & 0.648 & 0.636 & 0.634 & 0.641 \\
    Eq-4D-StOP \cite{zhu20234d} & 0.650 & 0.642 & 0.633 & 0.658 \\
    PTv3 \cite{wu2024point} & 0.662 & 0.659 & 0.627 & 0.670 \\
    4DSegStreamer (P3)  & 0.693 & 0.683 & 0.675 & 0.690 \\
    4DSegStreamer (M4F)  & \textbf{0.721} & \textbf{0.733} & \textbf{0.701} & \textbf{0.699} \\
    \bottomrule
  \end{tabular}
  \label{tab:ns_wopose}
\end{table}

\subsection{Streaming 4D Panoptic Segmentation in Outdoor datasets}
\nbf{SemanticKITTI \cite{behley2019semantickitti}}
Tab.~\ref{tab:sk_wopose} and \ref{tab:sk_wpose} compare streaming 4D panoptic segmentation on the SemanticKITTI validation split in the unknown and known pose settings. We compare our method with StreamYOLO \cite{yang2022real}, LongShortNet \cite{li2023longshortnet}, DAMO-StreamNet \cite{he2023damo}, Mask4Former \cite{yilmaz2024mask4former}, Eq-4D-StOP \cite{zhu20234d} and PTv3 \cite{wu2024point}. Originally designed for 2D streaming object detection via temporal feature fusion, the first three models are adapted to 4D streaming by replacing their backbones with P3Former \cite{xiao2024position}. Mask4Former and Eq-4D-StOP are designed for 4D panoptic segmentation but are not optimized for streaming. PTv3 is a state-of-the-art method designed for 3D perception. We adapt it to 4D panoptic segmentation with flow propagation according to \cite{aygun20214d}.

From both tables, we observe that 2D streaming methods perform poorly due to their reliance on real-time backbones, which are difficult to achieve in such a high-granularity task. Similarly, 4D panoptic segmentation methods also suffer significant performance degradation due to computational latency. PTv3 performs better than 4D methods due to its high efficiency, but it still suffers from performance drop. In contrast, our method outperforms all baseline models by a large margin in the streaming setting. Notably, in the unknown pose setting, our method achieves significant improvements of 7.7\% and 15.2\% in sLSTQ over PTv3 \cite{wu2024point}when integrated with P3Former and Mask4Former respectively, demonstrating the effectiveness of our alignment strategies across both dynamic and static classes. When combined with Mask4Former, our method outperforms its combination with P3Former, as Mask4Former is specifically designed for 4D panoptic segmentation.

\begin{table}
\caption{nuScenes validation set result in \textit{known pose} streaming setting. The best is highlighted in \textbf{bold}.}
  \centering
  \begin{tabular}{l|cccc}
    \toprule
    Method & sLSTQ & sPQ & sPQ$_{d}$ & sPQ$_{s}$ \\
    \midrule
    StreamYOLO \cite{yang2022real}  & 0.613 & 0.593 & 0.583 & 0.613 \\
    LongShortNet \cite{li2023longshortnet} & 0.628 & 0.6116 & 0.599 & 0.621 \\
    DAMO-StreamNet \cite{he2023damo}  & 0.633 & 0.625 & 0.607 & 0.639 \\
    Mask4Former \cite{yilmaz2024mask4former}  & 0.681 & 0.665 & 0.655 & 0.683 \\
    Eq-4D-StOP \cite{zhu20234d} & 0.695 & 0.673 & 0.654 & 0.693 \\
    4DSegStreamer (P3)  & 0.747 & 0.723 & 0.711 & 0.733 \\
    4DSegStreamer (M4F)  & \textbf{0.765} & \textbf{0.751} & \textbf{0.734} & \textbf{0.786} \\
    \bottomrule
  \end{tabular}
  \label{tab:ns_wpose}
\end{table}

\nbf{nuScenes \cite{caesar2020nuscenes}} 
We also compare the performance of 4D streaming panoptic segmentation on the nuScenes validation split \cite{caesar2020nuscenes}. Compared to SemanticKITTI\cite{behley2019semantickitti}, it has a slower frame rate, which allows many baseline methods to achieve real-time computation. However, in a streaming setting, even real-time methods experience at least a one-frame delay, leading to performance degradation. 
As shown in Tab.~\ref{tab:ns_wopose} and \ref{tab:ns_wpose}, our method outperforms all baseline approaches in both known and unknown pose settings. Additionally, all models perform better in the known pose setting, as pose estimation in the unknown pose setting takes more time, further degrading performance.

\subsection{Streaming 4D Panoptic Segmentation in Indoor dataset}
\indent 

\nbf{HOI4D \cite{liu2022hoi4d}} We also evaluate our model in indoor scenarios. We compare our approach with StreamYOLO \cite{yang2022real}, LongShortNet \cite{li2023longshortnet}, DAMO-StreamNet \cite{he2023damo}, NSM4D \cite{dong2023nsm4d} and PTv3 \cite{wu2024point}. As shown in Tab.~\ref{tab:hoi_wopose}, our method outperforms all other approaches, surpassing the runner-up by $6.6\%$ in terms of sLSTQ. This demonstrates that our method exhibits strong generalization ability, performing well not only in outdoor scenarios but also in indoor scenes.

\begin{table}
    \caption{HOI4D test set result in \textit{unknown pose} streaming setting. The best is highlighted in \textbf{bold}.}
  \centering
  \begin{tabular}{l|cccc}
    \toprule
    Method & sLSTQ & sPQ & sPQ$_{d}$ & sPQ$_{s}$ \\
    \midrule
    StreamYOLO \cite{yang2022real} & 0.373 & 0.336 & 0.362 & 0.324 \\
    LongShortNet \cite{li2023longshortnet} & 0.377 & 0.335 & 0.354 & 0.323 \\
    DAMO-StreamNet \cite{he2023damo} & 0.375 & 0.335 & 0.351 & 0.324 \\
    NSM4D \cite{dong2023nsm4d} & 0.314 & 0.305 & 0.315 & 0.303 \\
    PTv3 \cite{wu2024point} & 0.445 & 0.417 & 0.397 & 0.445 \\
    4DSegStreamer (P3) & 0.483 & 0.455 & 0.431 & 0.490 \\
    4DSegStreamer (M4F) & \textbf{0.511} & \textbf{0.482} & \textbf{0.457} & \textbf{0.533} \\
    \bottomrule
  \end{tabular}
  \label{tab:hoi_wopose}
\end{table}

\begin{table}
    \caption{General evaluation of different backbones. $w/o\ streamer$ is vanilla backbone. $w\ streamer$ is 3D or 4D backbone with our 4DSegStreamer.}
    \vspace{-0.5em}
  \centering
  \begin{tabular}{l|cc}
    \toprule
    Method & sLSTQ$_{w/o\ streamer}$ & sLSTQ$_{w\ streamer}$ \\
    \midrule
    Mask4Former \cite{yilmaz2024mask4former} & 0.515 & \textbf{0.688}   \\
    Eq-4D-StOP \cite{zhu20234d} & 0.504 & \textbf{0.674}  \\
    P3former \cite{xiao2024position} & 0.304 &  \textbf{0.613}  \\
    \bottomrule
  \end{tabular}
  \label{tab:general}
\end{table}

\begin{table}
    \caption{Ablation study in unknown pose streaming setting. $P3$ indicates the P3former backbone. $Mem$ represents the memory module. $Pose$ and $Flow$ denote multi-frames future pose and flow forecasting, respectively. $M\ Flow$ indicates the moving mask to assign non-zero flow only to moving objects.}
  \centering
  \begin{tabular}{l|ccc}
    \toprule
    Method & sLSTQ & sLSTQ$_{d}$ & sLSTQ$_{s}$ \\
    \midrule
    P3 \cite{xiao2024position}& 0.304 & 0.265 & 0.357\\
    P3+Mem & 0.349 & 0.292 & 0.408\\
    P3+Mem+Pose & 0.497 & 0.488 & 0.501\\
    P3+Mem+Pose+Flow & 0.591 & 0.667 & 0.514\\
    P3+Mem+Pose+M Flow & \textbf{0.613} & \textbf{0.682} & \textbf{0.516} \\
    \bottomrule
  \end{tabular}
  \label{tab:app_ab_unknown pose}
\end{table}

\subsection{Ablations for System}
\indent 

In this section, we conduct several groups of ablation studies on SemanticKITTI \cite{behley2019semantickitti} validation set to demonstrate the effectiveness of 4DSegStreamer.

\nbf{General to 3D and 4D backbone} Tab~\ref{tab:general} demonstrates that integrating our plug-and-play 4DSegStreamer consistently boosts the perfomance across various SOTA 3D and 4D backbones, with significnt improvements observed. This highlights the generality and effectiveness of our framework in enabling real-time capability.

\nbf{Effects of Components} Pose alignment mitigates the ego-pose motion, resulting in improvements to both $\text{sLSTQ}_d$ and $\text{sLSTQ}_s$. Building on this, incorporating flow alignment further refines the handling of moving objects, significantly boosting the model's performance on $\text{sLSTQ}_d$. We evaluate our method under both unknown-pose (Tab.~\ref{tab:app_ab_unknown pose}) and known-pose settings (Tab.~\ref{tab:app_ab_wpose}), where the latter provides ground-truth ego poses. Results demonstrate that our memory module, pose alignment, and dynamic object alignment continuously enhance streaming performance. Moreover, applying a non-moving object mask brings additional gains.


\nbf{Flow Forecasting Strategies}
We compare different flow forecasting strategies in Tab.~\ref{tab:ab_flow}. The "Inverse Forward Flow" represents a single iteration of the Inverse Flow Iteration algorithm, while the "Inverse Brute Search" algorithm directly searches for the forward flow within a restricted region that points to the target position. As shown in the table, forward flow forecasting does not achieve the best performance due to the high time consumption associated with repeated kd-tree construction. Additionally, backward flow forecasting performs poorly, as it is challenging to predict the backward flow without knowledge of the future position. In contrast, our proposed Inverse Flow Iteration algorithm shows superior performance in terms of sLSTQ.

\begin{table}
    \caption{Ablation study in known pose streaming setting. Pose is given and Flow is multi-head forecasting. $Mem$ represents the memory module. $Flow$ denotes multi-frame future flow forecasting.}
  \centering
  \begin{tabular}{l|ccc}
    \toprule
    Method & sLSTQ & sLSTQ$_{d}$ & sLSTQ$_{s}$ \\
    \midrule
    P3+Mem+GTpose & 0.563 & 0.534 & 0.592 \\
    P3+Mem+GTpose+Flow & \textbf{0.655} & \textbf{0.698} & \textbf{0.601}\\
    \bottomrule
  \end{tabular}
  \label{tab:app_ab_wpose}
\end{table}

\begin{table}
    \caption{Ablation study of different flow forecasting methods.}
  \centering
  \begin{tabular}{l|ccc}
    \toprule
    Method & sLSTQ & sLSTQ$_{d}$ & sLSTQ$_{s}$ \\
    \midrule
    Backward flow & 0.565 & 0.637 & 0.483 \\
    Forward flow & 0.589 & 0.667 & 0.497 \\
    Inverse forward flow & 0.586 & 0.662 & 0.502 \\
    Inverse brute search & 0.591 & 0.669 & 0.501 \\
    Inverse flow iteration & \textbf{0.613} & \textbf{0.682} & \textbf{0.516}\\
    \bottomrule
  \end{tabular}
  \label{tab:ab_flow}
\end{table}


\section{Conclusion}
\indent



In this work, we propose 4DSegStreamer, an efficient 4D streaming panoptic segmentation method that optimizes accuracy-latency trade-offs. We develop a dual-thread system to synchronize current and future point clouds within temporal constraints, complemented by an ego-pose forecaster and inverse forward flow iteration for motion alignment. Evaluated across diverse indoor and outdoor panoptic segmentation datasets, our method demonstrates robust performance in streaming scenarios.
\label{sec:conclusion}

{
    \small
    \bibliographystyle{ieeenat_fullname}
    \bibliography{main}
}

\appendix
\clearpage
\setcounter{page}{1}
\maketitlesupplementary
\renewcommand{\thetable}{S\arabic{table}}
\setcounter{table}{0} 
\renewcommand{\thefigure}{S\arabic{figure}}
\setcounter{figure}{0}



\section{Streaming Perception Setting}

\begin{figure}[t]
    \centering
    \includegraphics[width=1\linewidth]{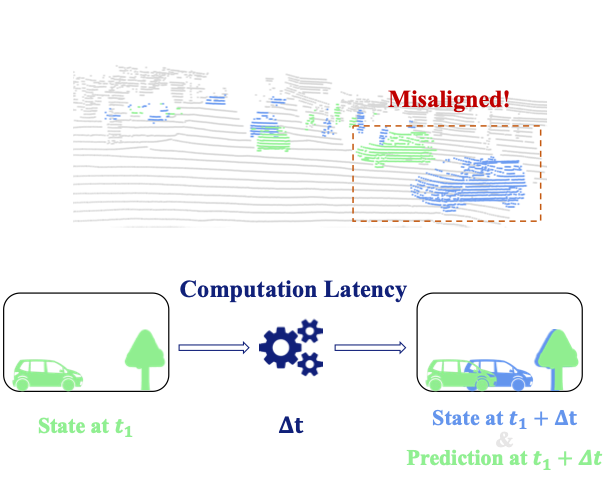}
    \caption{Streaming Perception Setting: \textcolor[rgb]{0.56,0.93,0.56}{Green} points denote dynamic objects from the processed frame, whereas \textcolor[rgb]{0.39,0.58,0.92}{blue} points represent the current frame at the time of prediction generated by the algorithm. }
    \label{fig:streaming_p_setting}
\end{figure}

Based on previous works \cite{li2020towards, yang2022real, he2023damo, li2023longshortnet, jo2022dade, huang2023mtd, zhang2024transtreaming, wang2023we} in streaming perception, our 4D streaming panoptic segmentation addresses a similar challenge by explicitly considering the impact of algorithmic processing latency on the final prediction and the scene at output time. As illustrated in Fig.~\ref{fig:streaming_p_setting}, predictions from existing methods are misaligned with the actual scene due to this latency. This misalignment can lead to perception inaccuracies, posing potential risks when robotic systems operate in highly dynamic environments.

\section{Forward Flow Iteration Proof}
\label{supp:proof}

To find the flow between the current query point and history position in geometric memory, we use the forward flow iteration. The iteration converges if Eq.~\ref{eq:converge} holds, then the following equation holds
\noindent
\begin{equation*}
\begin{aligned}
    1  \geq L & \geq \frac{\left| g(x_0+\Delta x) - g(x_0-\Delta x)  \right|}{\left| (x_0 + \Delta x) - (x_0 - \Delta x) \right|} \\ 
    & = \left| \frac{f(x_0+\Delta x) - f(x_0-\Delta x)}{(x_0 + \Delta x) - (x_0 - \Delta x)} \right|
    & = \left| f'(x_0) \right| 
\end{aligned}
\end{equation*}

For point $x$ on a rigid object and the flow $f(x,t)$ representing velocity, the derivative $\left|f^{\prime}(x) \right|$ can be expressed as:

\begin{equation*}
\begin{aligned}
    \left|f^{\prime}(x) \right| = \left|\frac{\partial f(x,t)}{\partial x} \right|& = \left|\frac{\partial \left(v + \omega \times (x-x_c)\right)}{\partial x}\right|\\
    & = \left|\frac{\partial (\omega \times x)}{\partial x} \right|= \left| \left[\omega\right]_{\times} \right|= \left| \omega\right|
\end{aligned}
\end{equation*}

where $x_c$ is the rotation center of the rigid body, $v$ is the translational velocity, $\omega$ is angular velocity, $\left[\omega\right]_{\times}$ is the cross-product matrix. The iteration converges when $\left|\omega\right| \leq 1$. In real-world scenarios, most rigid objects exhibit low angular velocity, allowing the iteration converges reliably. 

While perfect convergence cannot be guaranteed in practice, our experiments show robust convergence in 97.4\% of scenes in the SemanticKITTI dataset.

\section{Performance of different GPUs}
Table \ref{tab:gpu} presents the performance of our method across different GPUs under streaming settings. Since the model's runtime speed and GPU processing capability significantly impact the metric performance, the choice of hardware plays a crucial role. Notably, the A40 and 3090 graphics cards exhibit comparable performance due to their similar computational efficiency. In contrast, the A100 demonstrates a substantial speed advantage over the 3090, leading to a 1.4\% improvement in our model's performance on the A100.

\begin{table}
    \caption{Performance of different GPUs with different latency.}
    \vspace{-0.5em}
  \centering
  \begin{tabular}{l|ccc}
    \toprule
     & 4DSegStreamer(M4F) & PTv3 &  Mask4Former\\
    \midrule
    sLSTQ$_{A40}$ & 0.681 & 0.526 & 0.501\\
    sLSTQ$_{3090}$ & 0.688 & 0.536 & 0.504\\
    sLSTQ$_{A100}$ & 0.702 & 0.561 & 0.538\\
    \bottomrule
  \end{tabular}
  \label{tab:gpu}
\end{table}

\end{document}